\documentclass[sigconf]{acmart}
\AtBeginDocument{%
  }

\setcopyright{acmlicensed}
\copyrightyear{2026}
\acmYear{2026}
\acmDOI{XXXXXXX.XXXXXXX}
\acmConference[WWW Companion '26]{Proceedings of the ACM Web Conference 2026}{April 27--May 1, 2026}{Abu Dhabi, United Arab Emirates}
\acmISBN{978-1-4503-XXXX-X/26/04}

\usepackage{multirow}
\usepackage{graphicx}
\usepackage{subcaption}
\usepackage{booktabs}

\usepackage{tabularx}
\usepackage{booktabs}
\usepackage{balance}

\begin{document}
\title{TriFlow: A Progressive Multi-Agent Framework for Intelligent Trip Planning}

\author{Yuxing Chen}
\affiliation{%
  \institution{The University of Sydney}
  \city{Sydney}
  \country{Australia}
}
\email{yche0009@uni.sydney.edu.au}

\author{Basem Suleiman}
\authornote{Basem Suleiman is the corresponding author.}
\authornote{Basem Suleiman also with The University of Sydney}
\affiliation{%
  \institution{University of New South Wales}
  \city{Sydney}
  \country{Australia}
}
\email{b.suleiman@unsw.edu.au}

\author{Qifan Chen}
\affiliation{%
  \institution{The University of Sydney}
  \city{Sydney}
  \country{Australia}
}
\email{qifan.chen@sydney.edu.au}

\renewcommand{\shortauthors}{Chen et al.}

\begin{abstract}

Real-world trip planning requires transforming open-ended user requests into executable itineraries under strict spatial, temporal, and budgetary constraints while aligning with user preferences. Existing LLM-based agents struggle with constraint satisfaction, tool coordination, and efficiency, often producing infeasible or costly plans. 
To address these limitations, we present TriFlow, a progressive multi-agent framework that unifies structured reasoning and language-based flexibility through a three-stage pipeline of retrieval, planning, and governance. 
By this design, TriFlow progressively narrows the search space, assembles constraint-consistent itineraries via rule–LLM collaboration, and performs bounded iterative refinement to ensure global feasibility and personalisation. 
Evaluations on TravelPlanner and TripTailor benchmarks demonstrated state-of-the-art results, achieving 91.1\% and 97\% final pass rates, respectively, with over 10 $\times$ runtime efficiency improvement compared to current SOTA.

\end{abstract}




\maketitle

\section{Introduction}
Real-world trip planning is a complex task that combines natural language understanding, spatiotemporal constraint satisfaction, and multi-objective optimisation. Users often describe their travel goals in open-ended and ambiguous language
\begin{figure}[ht]
    \centering
    \includegraphics[width=0.9\linewidth]{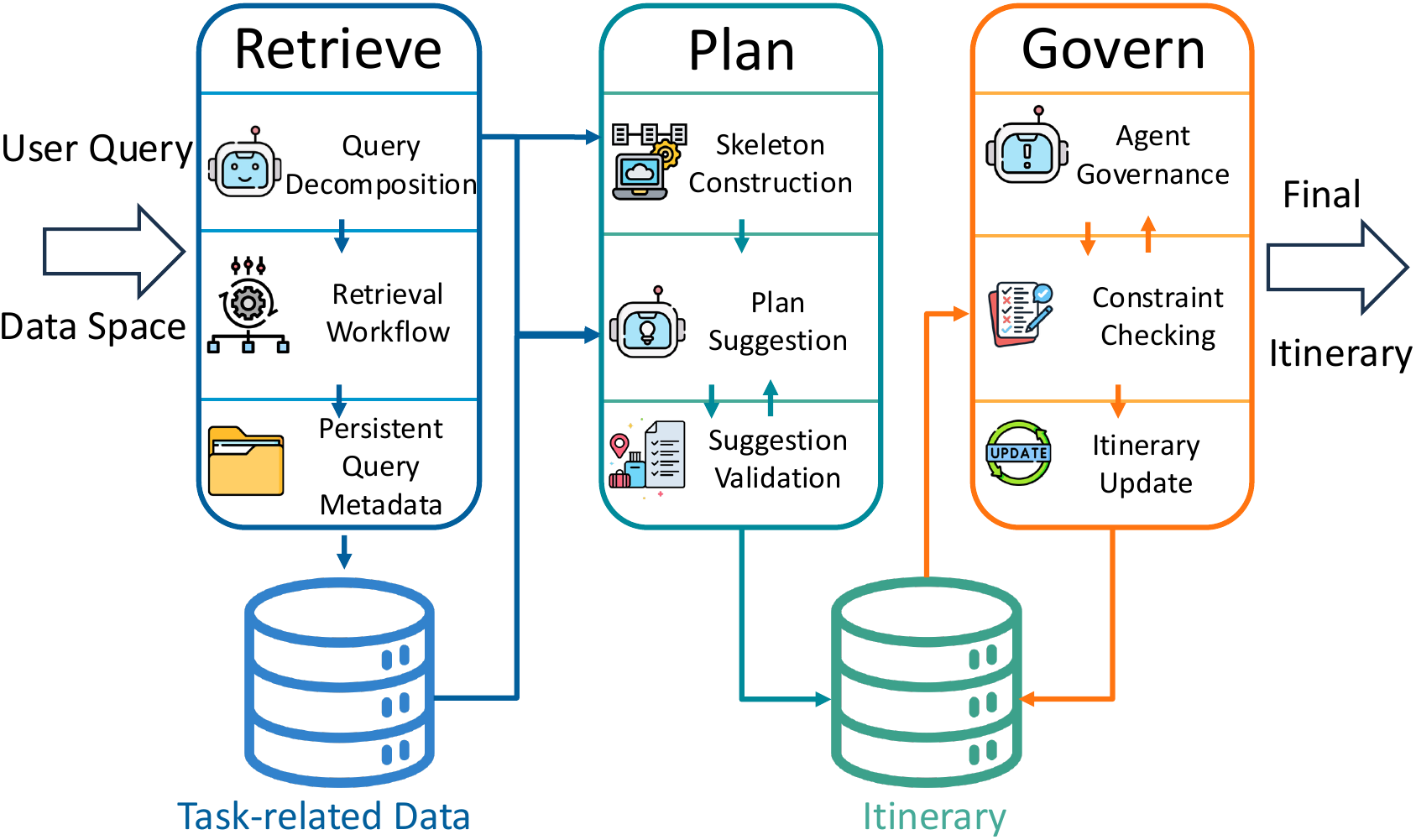}
    \caption[TriFlow Architecture Overview]{Three-stage progressive architecture of TriFlow.}
    \label{fig:tf_overview}
\end{figure}
(e.g., ``fun places,'' ``local food''), while executable itineraries must satisfy strict operational, temporal, spatial, and budgetary constraints. Despite rapid progress under the emerging ``LLM Agent + Tools'' paradigm, existing systems still struggle with real-world deployment. LLM-based agents that rely on tool generation often suffer from hallucinations, violate real-world constraints, and incur high token costs {\cite{xie2024travelplanner, wang2025triptailor}}. These limitations hinder both feasibility and personalisation.

Recent research has addressed these challenges through both benchmarking and methodological advances. New benchmarks such as TravelPlanner and TripTailor expand task scale, POI coverage, and verifiable metrics, exposing a persistent gap between hard-constraint satisfaction and human-level experiential quality~\cite{xie2024travelplanner, wang2025triptailor}. Methodologically, neuro-symbolic and optimisation-coupled frameworks translate natural-language intents into computable structures~\cite{hao2024large}, while multi-agent and ``generate–verify–retrieve'' systems enhance coordination and constraint checking through cooperative reasoning~\cite{gundawar2024robust}. Personalisation-oriented studies explore lightweight user modelling and preference elicitation to improve alignment~\cite{chen2409travelagent, jiang2024towards}. However, these works often evolve independently, lacking an integrated design that jointly ensures orchestration robustness and cost efficiency.
\begin{figure*}[ht]
    \centering
    \includegraphics[width=0.8\textwidth]{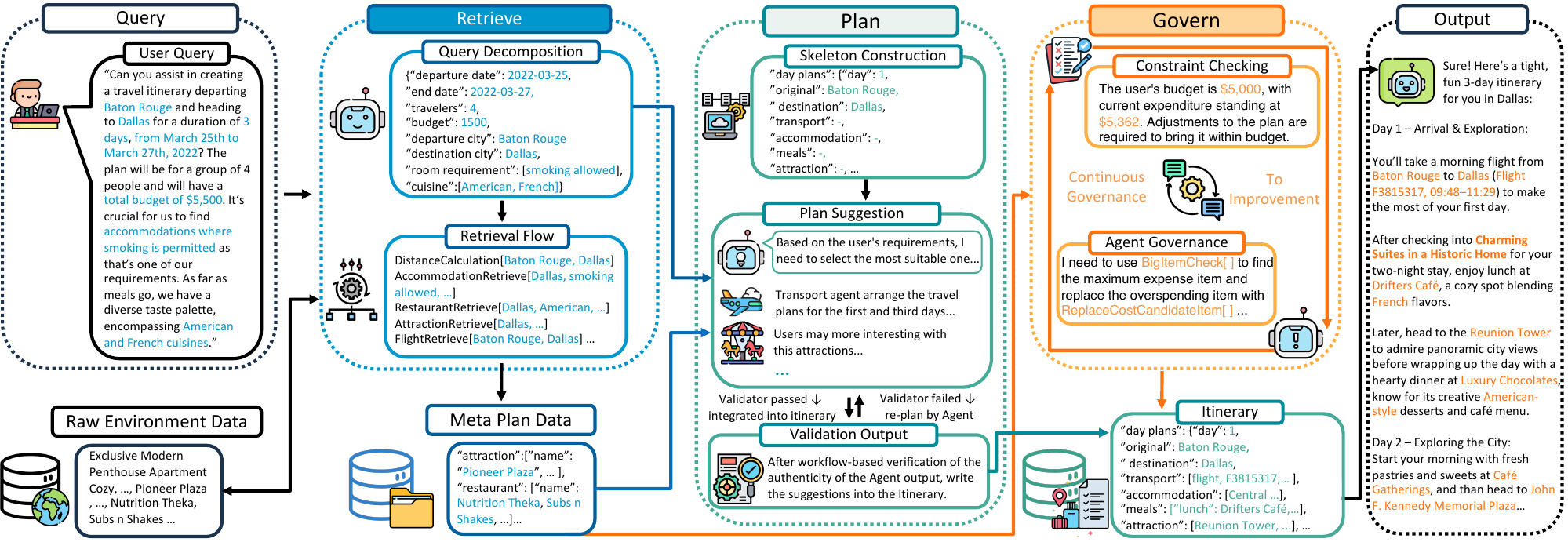}
    \caption[Demo Presentation of TriFlow]{Demonstration of how TriFlow takes a user’s natural query and passes it through structured retrieval, constraint-aware planning, and iterative governance, ultimately producing a feasible and personalised itinerary.}
    \label{fig:tf_main}
\end{figure*}
To bridge these gaps, we propose TriFlow, a progressive multi-agent framework for intelligent trip planning that unifies structured reasoning and flexible natural language understanding. TriFlow is built upon three principles: (1) a staged retrieval–planning–governance pipeline that progressively maps user requests into denoised structured intermediates, narrowing the solution space through explicit pruning; (2) a Rule–LLM collaborative division, where rule-based planning constructs a feasible skeleton and the LLM refines details within constraint boundaries; and (3) modular constraint orchestration that encapsulates temporal, spatial, budgetary, and preference conditions into pluggable modules managed by a state-machine controller. This design achieves reproducibility, interpretability, and cost efficiency while maintaining global feasibility.

Methodologically, TriFlow introduces a unified language-structure-constraint-iteration paradigm that aligns user intent with environmental data via observable intermediates and demand-driven pruning. Systemically, it provides a robust orchestration workflow that integrates modular constraint governance and conditional routing, forming a closed ``generate–verify–assemble–recompute'' loop. Empirically, experiments on public trip-planning benchmarks demonstrated consistent improvements in feasibility, efficiency, and personalisation over prior LLM-agent systems.

In summary, our work makes three key contributions. First, we introduce TriFlow, a progressive multi-agent framework for intelligent trip planning. Second, we conduct extensive experiments on two authoritative benchmarks, demonstrating the effectiveness and robustness of TriFlow. Finally, we analyse the key design factors that explain TriFlow’s strong performance.

\section{Methodology}
TriFlow formulates trip planning as a feasibility-first optimisation problem that progressively refines natural-language inputs into executable itineraries under real-world constraints. Unlike conventional LLM-based planners that optimise for textual fluency or coverage, TriFlow embeds constraint satisfaction (temporal, spatial, budgetary, and preferential) as the primary decision layer, ensuring that all optimisation steps operate strictly within a feasible domain.

\textbf{\textit{Problem formulation.}}
A user request contains information such as destination cities, travel dates, number of travellers, budget, and personal preferences. The global data space includes flights, accommodations, points of interest, distances, and other factual resources. Based on the request, TriFlow retrieves a task-specific subset of relevant data and incrementally synthesises an itinerary that satisfies temporal, spatial, and budget constraints while aligning with user preferences. The final refined itinerary is the output itinerary.
 
\begin{figure*}[ht]
\centering
\begin{subfigure}{0.34\textwidth}
    \includegraphics[width=\linewidth]{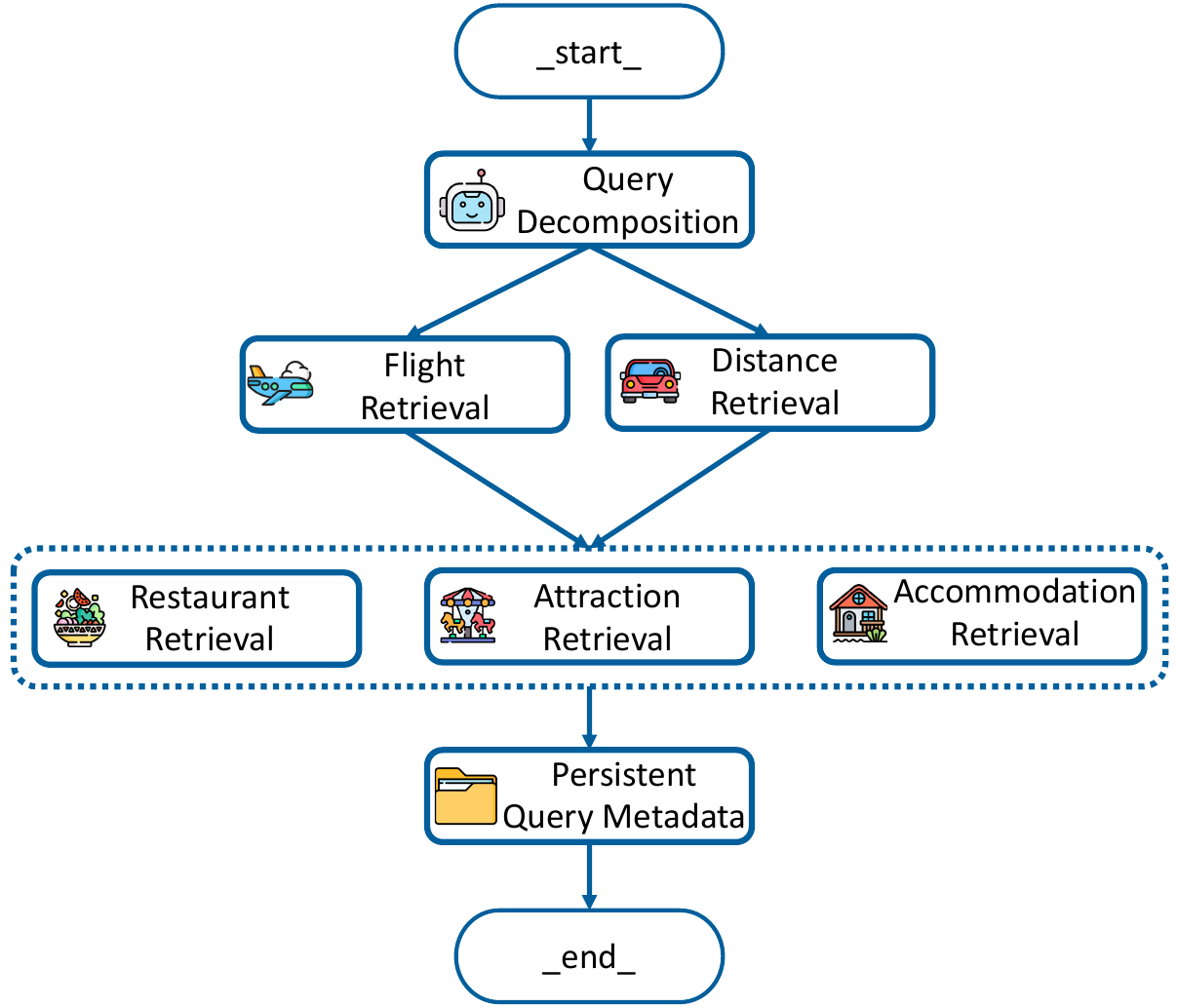}
    \caption{Retrieval Stage}
    \label{fig:s1}
\end{subfigure}
\hspace{0.008\textwidth}
\begin{subfigure}{0.27\textwidth}
    \includegraphics[width=\linewidth]{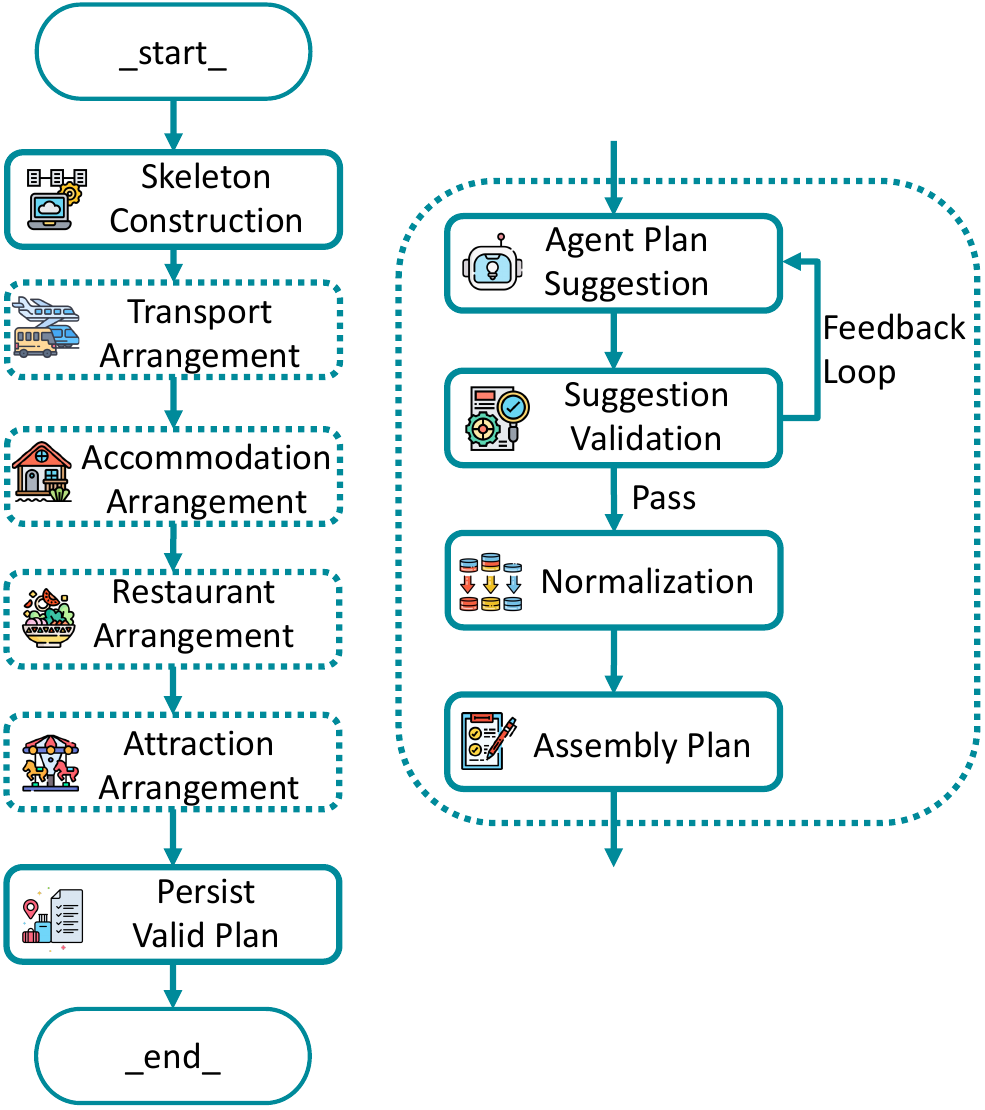}
    \caption{Planning Stage}
    \label{fig:s2}
\end{subfigure}
\hspace{0.008\textwidth}
\begin{subfigure}{0.31\textwidth}
    \includegraphics[width=\linewidth]{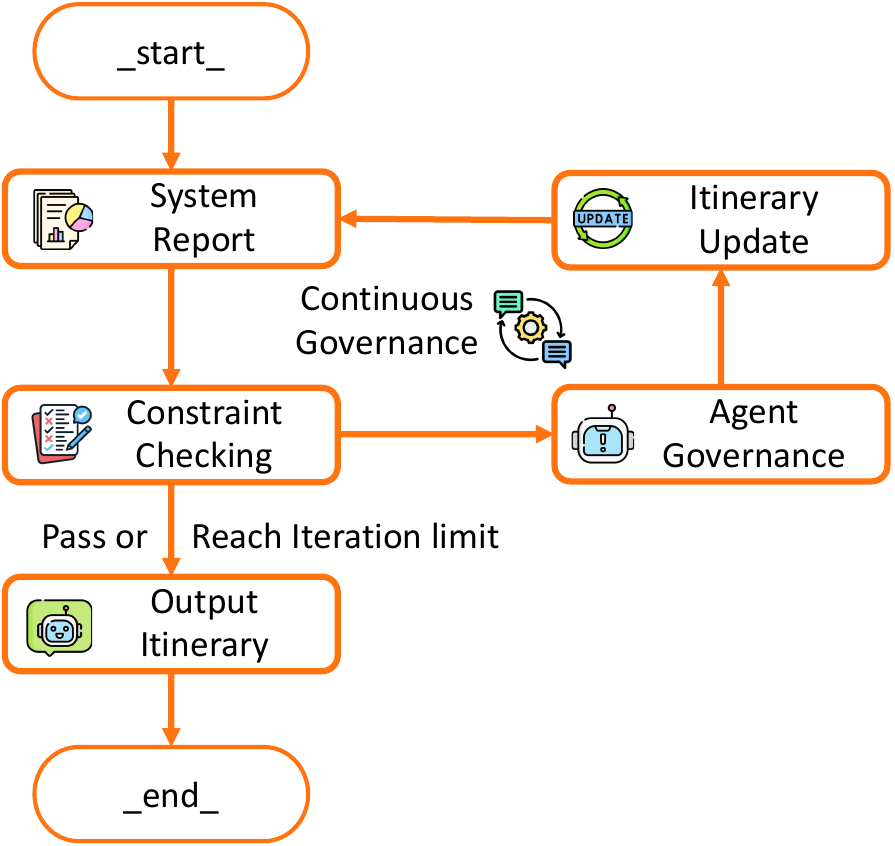}
    \caption{Governance Stage}
    \label{fig:s3}
\end{subfigure}
\caption[TriFlow Workflow Overview]{Overview of the TriFlow framework: each stage progressively contracts the search space}
\label{fig:triflow_overview}
\end{figure*}

TriFlow performs optimisation in a feasibility-first hierarchy. Each stage progressively narrows the feasible space while preserving overall validity. Rather than directly optimising raw text, TriFlow operates within structured representations and improves itinerary quality only after all constraints are met. This process is organised into three progressively constrained stages: Retrieval, Planning, and Governance, as shown in Figure~\ref{fig:triflow_overview}. Each stage is responsible for a single type of operation, such as information recall, structural assembly, or controlled refinement, and all stages are equipped with explicit feasibility checks. This decomposition improves interpretability, fault isolation, and scalability compared with monolithic LLM planners.

\subsection{System Architecture}
\hspace*{1em}\textbf{\textit{Stage I: Retrieval.}}
Given a user request and the global data space, the retrieval stage produces a task-specific subset of factual resources that defines the boundary for downstream planning. LLM agents first perform query decomposition to translate open-ended requests into structured requirements. Parallel modules then retrieve relevant entities such as flights, distances, restaurants, attractions, and accommodations, followed by validation and deduplication. Integrity checks for geometry, time windows, and price consistency ensure that the retrieved subset remains coherent and reliable.

\textbf{\textit{Stage II: Planning.}}
The planning stage constructs a feasible itinerary within the retrieved factual space. It incrementally builds a coarse-to-fine structure. This process begins by determining the order of cities and allocating time across destinations, and continues with filling in details for transportation, lodging, meals, and activities. These steps are completed through agent–validator loops that consist of suggestion, validation, and normalisation. User preferences are incorporated through interpretable arbitration, which ranks feasible candidates by alignment and efficiency. All modules adhere to a monotonic feasibility principle: once a constraint is satisfied, subsequent steps are not allowed to violate it. For instance, once the city's order and daily arrangement windows are fixed, later steps (e.g., selecting restaurants or attractions) must operate within these bounds rather than revising earlier structural decisions.

\textbf{\textit{Stage III: Governance.}}
The governance stage refines the feasible itinerary through bounded iterative improvements. Each iteration begins with a system report that summarises budget usage, timing consistency, and preference satisfaction, followed by constraint checking to identify any violations. When issues or opportunities for improvement are detected, the governance agent proposes targeted adjustments, such as replacing costly items, resolving timing conflicts, or improving alignment with user preferences. The itinerary is updated only when the proposed changes preserve feasibility and provide clear benefits. The refinement loop is capped at a small fixed number of iterations and terminates early if the itinerary converges or no further feasible improvements can be made. This process ensures that the itinerary stabilises into a fully feasible and high-quality plan.

Together, these three stages embody TriFlow's philosophy of feasibility-first. The retrieval stage bounds the factual domain, the planning stage assembles valid structures, and the governance stage ensures refinement and consistency. Through progressive optimisation and coordinated interaction between rules and LLM agents, TriFlow transforms ambiguous natural-language requests into executable and personalised itineraries with guaranteed feasibility and efficiency.

\section{Results}
\label{sec:results}
\hspace*{1em}\textbf{\textit{Benchmarks selection.}}
We evaluated TriFlow on two authoritative benchmarks, TravelPlanner (validation set) and TripTailor\cite{xie2024travelplanner, wang2025triptailor}, strictly followed the official evaluation protocols and scripted checkers of each benchmark, reporting Delivery Rate, Commonsense and Hard-constraint pass rates, Feasibility and Rationality pass rates, and the Final Pass Rate (FPR).

\textbf{\textit{Configuration.}}
To ensure fair comparison with existing SOTA on benchmarks, the TravelPlanner experiment uses GPT-4o, and the TripTailor setting uses GPT-4o-mini as a base model. TriFlow applies a staged temperature schedule across its three phases (retrieval = 0.0, planning = 0.3, governance = 0.6) to balance determinism and creativity. The governance stage is further limited to a maximum of eight refinement iterations.

\subsection{TravelPlanner}
\label{sec:travelplanner_results}
Table~\ref{tab:main-res} summarises the validation results on 180 instances of the TravelPlanner benchmark. 
TriFlow achieved a 100\% Delivery Rate, 95.0\% macro commonsense and 96.1\% macro hard-constraint pass rates, yielding a 91.1\% FPR.

This performance matched or slightly surpassed the current state-of-the-art (SOTA) FormalVerify. 
TriFlow completed each task in 22.6s on average, over 10.9$\times$ faster than FormalVerify (245.7s), demonstrating strong runtime efficiency in real-time trip synthesis. 
Across all eight commonsense and five hard-constraint categories, TriFlow exceeded 95\% pass rates and maintained stable performance across Easy/Medium/Hard tiers (96.7/95.0/80.0\% FPR), indicating globally consistent and constraint-compliant itineraries under long-horizon planning conditions.
\subsection{TripTailor}
\label{sec:triptailor_results}
Table~\ref{tab:main-triptailor} reports results on the TripTailor benchmark, validating TriFlow’s generalisation to large-scale real-world data. TriFlow achieved 99.1\% macro feasibility, 97.7\% macro rationality, and a 97.7\% FPR, outperforming the strongest workflow baseline by more than 34\%.

Performance remained robust across difficulty splits (Easy/Hard macro feasibility 99.4/98.9\%, macro rationality 98.3/97.1\%). 
TriFlow also achieved achieving $\geq$97.6\% on all metrics across all feasibility and rationality dimensions, including Within Sandbox (99.1\%), Complete Information (99.9\%), restaurant and attraction diversity ($\ge$99.7\%), budget compliance (99.9\%), and duration consistency (97.6\%), confirming its ability to produce coherent, high-quality itineraries in realistic environments.

Across both benchmarks, TriFlow consistently delivered feasible-first plans with strong rationality while reducing latency by an order of magnitude relative to formal verification pipelines. The gains aligned with our staged design: retrieval constrained the factual domain, planning assembled a valid structure with validator-guarded choices, and governance performed bounded, explainable refinement. 
\begin{table}[ht]
    \centering
    \caption{Main results on TravelPlanner.}

\begin{tabular*}{\columnwidth}{@{\extracolsep{\fill}}lccccc}
\toprule
\multirow{2}{*}{\centering\textbf{Method}}
& \multicolumn{2}{c}{\begin{tabular}[c]{@{}c@{}}Commonsense\end{tabular}} 
& \multicolumn{2}{c}{\begin{tabular}[c]{@{}c@{}}Hard Constraint\end{tabular}} 
& \multirow{2}{*}{\begin{tabular}[c]{@{}c@{}}FPR\end{tabular}} \\ 
\cmidrule(l){2-3} \cmidrule(l){4-5}
& Micro & Macro & Micro & Macro & \\ 
\midrule

TravelPlanner\cite{xie2024travelplanner} & 80.4 & 17.2 & 47.1 & 22.2 & 4.4 \\

FormalVerify\cite{hao2024large} & 95.0 & \textbf{95.0} & 95.7 & \textbf{98.9} & \textbf{93.3} \\
TriFlow (ours) & \textbf{99.3} & \textbf{95.0} & \textbf{99.2} & 96.1 & 91.1 \\
\bottomrule
\end{tabular*}
    \label{tab:main-res}
\end{table}
\begin{table}[ht]
\centering
\caption{Main results on TripTailor.}

\begin{tabular*}{\columnwidth}{@{\extracolsep{\fill}}lccccc}
\toprule
\multirow{2}{*}{\centering\textbf{Method}}
& \multicolumn{2}{c}{\begin{tabular}[c]{@{}c@{}}Feasibility\end{tabular}} 
& \multicolumn{2}{c}{\begin{tabular}[c]{@{}c@{}}Rationality\end{tabular}}
& \multirow{2}{*}{\begin{tabular}[c]{@{}c@{}}FPR\end{tabular}} \\ 
\cmidrule(l){2-3} \cmidrule(l){4-5}
& Micro & Macro & Micro & Macro & \\ 
\midrule
Workflow\cite{wang2025triptailor} & 98.3 & 97.3 & 91.6 & 63.7 & 63.3 \\
Direct\cite{wang2025triptailor} & 98.3 & 96.6 & 76.7 & 22.6 & 21.5 \\
TriFlow (ours) & \textbf{99.5} & \textbf{99.1} & \textbf{99.4} & \textbf{97.7} & \textbf{97.7} \\
\bottomrule
\end{tabular*}

\label{tab:main-triptailor}
\end{table}

\section{Discussion}

Table~\ref{tab:cspr_tavelplanner} summarised constraint-level performance across different task difficulties, revealing where TriFlow’s staged, feasibility-first design yielded the greatest impact. 
The most pronounced improvement was seen in the Final Pass Rate, which rose from near-zero to over 80\% under challenging conditions, confirming that TriFlow consistently delivers executable itineraries even as constraint density increases.

TriFlow’s retrieval stage substantially enhanced factual consistency, with metrics such as Within Sandbox improving from about 33–50\% to over 95–100\%, demonstrating that query decomposition and validation effectively eliminated irrelevant or incoherent data. This bounding of the factual domain formed the foundation for downstream feasibility.

During itinerary assembly, TriFlow’s agent–validator loop enforced monotonic feasibility; once a constraint was satisfied, it could not be violated later. 
This yielded sharp gains in structural constraints such as Reasonable City Route, Within Current City, and Minimum Nights Stay, which all approached or exceeded 98\%. 
These improvements validated the effectiveness of skeleton-first construction and structured assembly in preserving temporal–spatial coherence.

TriFlow also have a significant improvement on hard constraints. The reason is that the planning stage adheres to a monotonic feasibility principle, which establishes the core structural decisions early, leaving only a small number of residual violations that cannot be corrected without violating earlier commitments. Governance then applies bounded iterative refinement to resolve these remaining issues, particularly for highly entangled constraints. As a result, metrics such as Budget increased from 4–10\% to over 95\%, and other hard constraints all achieved pass rates above 95\%.

Collectively, these item-level improvements substantiated the rationale of TriFlow’s staged design: 
\begin{table}[ht]
\caption{Constraint pass rate on TravelPlanner (TravelPlanner/TriFlow).}
\centering

\begin{tabularx}{\columnwidth}{l *{3}{>{\centering\arraybackslash}X}}
\toprule
\textbf{Constraint Type} & \textbf{Easy} & \textbf{Middle} & \textbf{Hard} \\
\midrule
\multicolumn{4}{c}{\textbf{\textit{Commonsense Constraint}}} \\
\midrule
Within Sandbox           & 37.4/100 & 31.2/95.0 & 33.9/98.3 \\
Complete Information     & 53.4/98.3 & 52.9/100 & 58.0/98.3 \\
Within Current City      & 69.3/100 & 67.3/100 & 68.3/100 \\
Reasonable City Route    & 44.5/98.3 & 45.6/100 & 54.9/96.7 \\
Diverse Restaurants      & 85.1/100 & 81.4/100 & 86.8/100 \\
Diverse Attractions      & 94.3/100 & 90.4/100 & 94.0/100 \\
Non-conf. Transportation & 70.1/100 & 73.3/100 & 83.1/100 \\
Minimum Nights Stay      & 46.8/100 & 46.2/100 & 51.1/98.3 \\
\midrule
\multicolumn{4}{c}{\textbf{\textit{Hard Constraint}}} \\
\midrule
Budget                  & 10.1/100 & 8.4/100 & 4.4/95.0 \\
Room Rule               & --/-- & 5.6/100 & 11.3/98.3 \\
Cuisine                 & --/-- & 10.8/100 & 11.4/96.7 \\
Room Type               & --/-- & 12.4/100 & 13.8/100 \\
Transportation          & --/-- & --/-- & 18.6/100 \\
\midrule
\multicolumn{4}{c}{\textbf{\textit{Final}}} \\
\midrule
Final Pass Rate         & \textbf{1.1/96.7} & \textbf{0.3/95.0} & \textbf{0.3/80.0} \\
\bottomrule
\end{tabularx}

\label{tab:cspr_tavelplanner}
\end{table}
retrieval eliminated inconsistent data, planning guaranteed structural feasibility, and governance enforced bounded optimisation of hard constraints. 
The framework’s progressive narrowing of feasible domains translates directly into empirical reliability, providing interpretable, high-quality trip plans even under challenging real-world conditions.

\section{Conclusion}
TriFlow establishes a feasibility-first paradigm for intelligent trip planning, unifying retrieval, planning, and governance through structured reasoning and LLM-based generation. The system progressively transforms open-ended natural-language requests into executable itineraries that satisfy real-world constraints while remaining interpretable and personalised. Experiments on two public benchmarks demonstrate that TriFlow consistently achieves high feasibility and rationality, with competitive runtime efficiency, underscoring its promise as a practical foundation for reliable LLM-driven planning. Future work will extend TriFlow from offline benchmark evaluations to real-time online environments, enabling live data retrieval and testing its robustness under dynamic, real-world conditions.

\balance
\bibliographystyle{ACM-Reference-Format}
\bibliography{mian}

\end{document}